\documentclass{article} %
\usepackage{iclr2027_conference,times}
\usepackage[T1]{fontenc}
\usepackage[varqu]{zi4}

\usepackage{amsmath,amsfonts,bm}

\def\eqref#1{equation~\ref{#1}}

\def\1{\bm{1}}

\DeclareMathAlphabet{\mathsfit}{\encodingdefault}{\sfdefault}{m}{sl}
\SetMathAlphabet{\mathsfit}{bold}{\encodingdefault}{\sfdefault}{bx}{n}

\usepackage{hyperref}
\usepackage{url}
\usepackage{amsmath,amssymb}
\usepackage{booktabs}
\usepackage{array}
\usepackage{graphicx}
\usepackage{wrapfig}
\usepackage{fancyvrb}
\usepackage{placeins}   %
\usepackage{flafter}   %
\usepackage{xcolor}
\newif\ifdraft \drafttrue

\newcommand{\method}{Attention Relay}

\definecolor{navy}{rgb}{0,0,0.5}   %
\definecolor{gainblue}{HTML}{2A78D6}   %
\hypersetup{colorlinks=true, linkcolor=navy, citecolor=navy, urlcolor=navy, filecolor=navy}
\hypersetup{pdftitle={Lend Me Your Eyes: Instruction-Aware Text Embeddings via Attention Relay}, pdfauthor={Yiyuan Luo, Vaggos Chatziafratis}}

\title{Lend Me Your Eyes: Instruction-Aware \\ Text Embeddings via Attention Relay}

\iclrfinalcopy
\author{Yiyuan Luo \\
University of California, Santa Cruz \\
\texttt{yluo124@ucsc.edu} \\
\And
Vaggos Chatziafratis \\
University of California, Santa Cruz \\
\texttt{vaggos@ucsc.edu}
}

\begin{document}

\maketitle
\lhead{Preprint}   %
\suppressfloats[t]   %

\begin{abstract}
Text embedding models trained with contrastive learning learn to follow task instructions from instruction-paired data, while instruction-tuned LLMs already know how to follow them. We show that this instruction-following ability can carry over from an LLM to a Transformer-based embedder without any training. We propose Attention Relay, which passes the attention weights an LLM produces to the embedder's own attention. Across six instruction-tuned LLMs from the Qwen3, Llama 3.1 and OLMo 3 families and ten widely used embedding models that differ in tokenizer, size and pooling type, Attention Relay makes nearly every combination instruction-aware. Experiments that break the method down into its parts show that the LLM's attention weights track the instruction in its later layers and come largely from instruction tuning. They also show that relaying these weights selects which content in the text matters: it makes the aspect of the text that the instruction asks about dominant in the embedding, or restores that aspect where averaging had diluted it.

\smallskip
\begin{tabular}{@{}ll@{}}
Project Page: & \href{https://qdd99.github.io/attention-relay}{\nolinkurl{qdd99.github.io/attention-relay}} \\
Code: & \href{https://github.com/qdd99/attention-relay}{\nolinkurl{github.com/qdd99/attention-relay}}
\end{tabular}

\end{abstract}

\section{Introduction}\label{sec:intro}

Text embeddings are central to search, retrieval-augmented generation, clustering and classification \citep{muennighoff2023mteb}, and the success of Large Language Models (LLMs) has made them stronger: the best embedding models are now built on LLM backbones \citep{wang2024improving,behnamghader2024llm2vec, muennighoff2024gritlm,zhang2025qwen3embedding}.
What makes two texts similar, however, can depend on the task: news articles can be grouped by topic or by where they happened, customer messages by how the customer feels or by what the customer needs.
We call any model or readout that turns a text into one vector by pooling its token states an \emph{embedder}; trained embedding models are the most common kind.

An embedder trained without instructions cannot follow one, even when the instruction is placed in its input.
Most recent embedders are therefore trained with instructions: they take an instruction as part of their input and learn to use it by contrastive training on texts paired with instructions, from InstructOR \citep{su2023instructor} to Qwen3-Embedding \citep{zhang2025qwen3embedding}, which takes curated instruction data and a round of training for each embedder, to teach what the LLMs they are built on already do \citep{wei2022flan,ouyang2022instructgpt}.

An instruction-tuned LLM follows such instructions readily, but it is no substitute for an embedder.
Its hidden states were never shaped for similarity, and embeddings read from them without training, such as PromptEOL \citep{jiang2024scaling}, Echo embeddings \citep{springer2024repetition} and Value Aggregation \citep{zhang2026values}, remain well behind trained embedders.
Nor does the LLM's grasp of an instruction obviously carry over to an embedder, which differs from it in size, tokenizer and representation space.

\begin{figure}[t]
\centering
\includegraphics[width=\linewidth]{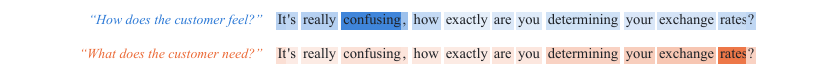}
\caption{One customer message read by Qwen3-1.7B under two questions, each token shaded by its reading weight.}
\label{fig:intro}
\end{figure}

We show that one part of an instruction-tuned LLM's instruction following does carry over, without any training: where the LLM reads a text when it is about to answer, the attention it pays to the text's tokens, which we call its \emph{reading weights} (Figure~\ref{fig:intro}).
We propose \emph{\method{}}, which hands these weights to the pooling of an instruction-agnostic embedder.
The embedder never sees the instruction, and the embedding stays in the space the embedder was trained for.

We apply \method{} to instruction-tuned LLMs of the Qwen3, Llama~3.1 and OLMo~3 families \citep{yang2025qwen3,grattafiori2024llama,olmo2025olmo3} and to widely used embedders, from the sentence-transformers models all-MiniLM-L6-v2 and all-mpnet-base-v2 \citep{reimers2019sbert} to BGE, GTE and E5 \citep{xiao2024cpack,li2023gte,wang2022e5} and Qwen3-Embedding \citep{zhang2025qwen3embedding}, which differ in tokenizer, size and kind of pooling. We show that \method{} makes the embedders instruction-aware in nearly every combination of LLMs and embedders, on the New York Times (NYT) and IntentEmotion datasets and on InBedder's other sets \citep{peng2024inbedder}.

\section{Motivation}\label{sec:motivation}

In this section, we show that an instruction-agnostic embedder holds what an instruction asks about but reads every text the same way, while an instruction-tuned LLM reads a text differently as the question changes.

\subsection{An embedder holds what an instruction asks about but cannot weight it}\label{sec:motivation-embedder}

\begin{figure}[t]
\centering
\begin{minipage}[t]{0.43\linewidth}
\centering
\includegraphics[width=\linewidth]{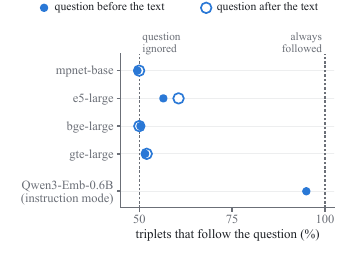}
\caption{Question following in embedders without vs.\ with instruction training.}
\label{fig:question-input}
\end{minipage}\hfill
\begin{minipage}[t]{0.54\linewidth}
\centering
\includegraphics[width=\linewidth]{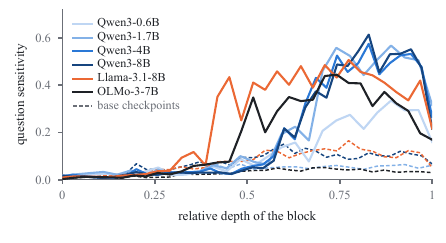}
\caption{Question sensitivity at every block of each LLM (solid) and its base checkpoint (dashed).}
\label{fig:premises}
\end{minipage}
\end{figure}

An embedding is a weighted average over a text's tokens.
Mean pooling gives every token the same weight, and CLS or last-token pooling lets one token attend to the others with attention the embedder learned for similarity in general.
In an embedder trained without instructions, nothing in these weights responds to a question, and a question placed in the input is only more text.
Asked whether a customer message is closer to one that shares its emotion or one that shares its intent, such an embedder follows the question in about half of the cases, as often as one that ignores it, while Qwen3-Embedding, trained with instructions, follows it in nearly all (Figure~\ref{fig:question-input}).
A news article about a business deal in Japan belongs with business news when articles are grouped by topic and with news from Japan when they are grouped by country, yet such an embedder gives it one vector for both, although generic embeddings do encode what an instruction asks about \citep[GSTransform;][]{feng2025gstransform}.
The embedder holds what an instruction asks about but has no way to weight it by the instruction.

\subsection{An instruction-tuned LLM reads the text for the question}\label{sec:motivation-llm}

An instruction-tuned LLM given a text followed by a question computes every state of the text before it sees the question.
Because of the causal mask, the text's states are the same under every question, and the question can act on the text only through where later positions attend.
At the position where the LLM is about to answer, its attention over the text, summed over heads and renormalized over the text, is what we call its reading weights (Figure~\ref{fig:intro}).

We measure how much the reading weights move with the question as the total variation distance between them under two questions about the same text, $\tfrac{1}{2} \sum_j |w_j(x, q_1) - w_j(x, q_2)|$, averaged over news articles asked for their topic or their location and customer messages asked for the customer's feeling or need.
In every instruction-tuned LLM we examine, this \emph{question sensitivity} stays near zero through the first half of the network, rises between 40 and 70\% of its depth, and falls again at the last block (Figure~\ref{fig:premises}).
Instruction tuning largely creates this reading: the base checkpoints' weights move far less with the question at every block.

As noted in Section \ref{sec:intro}, what the LLM lacks is a representation made for similarity: its hidden states were never shaped for it. The two models therefore complement each other: the LLM knows which parts of a text matter for the question, and the embedder knows how to represent them.

\section{\method{}}\label{sec:method}

\begin{figure}[t]
\centering
\includegraphics[width=\linewidth]{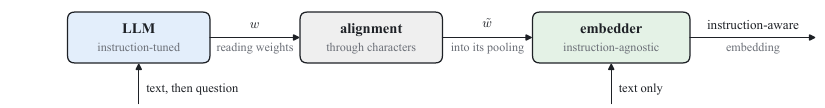}
\caption{\method{}: the LLM's reading weights, carried through characters to the embedder's tokens, pool the embedder's own states.}
\label{fig:method}
\end{figure}

\method{} makes an instruction-agnostic embedder instruction-aware by giving it, for each text, the weights with which the LLM reads that text under the instruction (Figure~\ref{fig:method}).

\subsection{Formulation}\label{sec:relay}

\paragraph{Reading weights.}
Let $x = (x_1, \dots, x_n)$ be a text in the LLM's tokens and $q$ an instruction.
We give the LLM the text followed by the instruction in its chat template (Appendix~\ref{app:prompt}) and take the last position of the prompt, where the model is about to answer, which we call the \emph{answer position}.
As in PromptEOL \citep{jiang2024scaling}, the prompt asks for a one-word answer, so that the token the model is about to write is the answer itself rather than a lead-in such as ``The''.
At block $\ell$, let $\alpha^{h}_{j}(x, q)$ be the attention from the answer position to text token $x_j$ in head $h$.
The LLM's \emph{reading weights} are these attentions summed over the block's $H$ heads and renormalized over the text:
\begin{equation}\label{eq:reading}
w_j(x, q) \;=\; \frac{\sum_{h=1}^{H} \alpha^{h}_{j}(x, q)}{\sum_{k=1}^{n} \sum_{h=1}^{H} \alpha^{h}_{k}(x, q)}, \qquad j = 1, \dots, n.
\end{equation}
Because the text precedes the instruction, the causal mask keeps the text's states identical under every instruction; the instruction changes only where the answer position looks.
The block $\ell$ is set once per LLM and is the same for every embedder and dataset.

\paragraph{From the LLM's tokens to the embedder's.}
The embedder splits $x$ into its own tokens, which need not match the LLM's.
We carry $w$ over through characters: each LLM token's weight is spread evenly over its non-space characters, and each embedder token $i$ receives the weight of the characters it covers,
\begin{equation}\label{eq:align}
\tilde{w}_i \;\propto\; \sum_{j=1}^{n} w_j\, \frac{|C_i \cap C_j|}{|C_j|},
\end{equation}
where $C_j$ and $C_i$ are the non-space characters of LLM token $j$ and embedder token $i$, and $\tilde{w}$ is renormalized over the embedder's text tokens.
Tokens outside the text, such as special tokens or a fixed prefix, cover none of its characters and receive no weight.

\paragraph{Relaying the weights.}
Let $r$ be the weights that the embedder's pooling places on its tokens.
In mean pooling, $r$ is uniform over the tokens and averages their final states.
In CLS or last-token pooling, the embedding is the final state of one pooling token; at each layer, a head's attention row $a$ from that token gives $r_i = a_i / m$ on the text tokens $i \in \mathcal{T}$, where $m = \sum_{i \in \mathcal{T}} a_i$ is the head's total attention on the text.
\method{} multiplies $r$ by the LLM's weights and renormalizes,
\begin{equation}\label{eq:relay}
r'_i \;\propto\; r_i \, \tilde{w}_i^{\,\beta},
\end{equation}
with strength $\beta = 1$ unless stated otherwise; $\beta = 0$ returns the released embedder, and a larger $\beta$ concentrates $r'$ on the tokens the LLM weights most.
The product keeps the embedder's own preferences among the tokens the LLM reads and removes weight from the tokens it ignores.
In mean pooling, $r'$ becomes the pooling weights.
In CLS or last-token pooling, every head at every layer spends its text attention $m$ as $m\,r'$, keeps its attention outside the text,\footnote{In a causal embedder the first token serves as an attention sink \citep{xiao2024streamingllm}; we count it outside the text, so its attention is kept as well.} and recomputes its output at the pooling token from the embedder's own values.

\paragraph{An equivalent logit form.}
In CLS or last-token pooling, the same change can be made before the softmax.
Let $s$ be the logits of a head's attention row at the pooling token, so that $a = \operatorname{softmax}(s)$.
For any bias $b$ over the positions,
\begin{equation}\label{eq:shift}
\operatorname{softmax}(s + b)_i \;=\; \frac{e^{s_i}\, e^{b_i}}{\sum_k e^{s_k}\, e^{b_k}} \;=\; \frac{a_i\, e^{b_i}}{\sum_k a_k\, e^{b_k}},
\end{equation}
so adding $b_i$ to a logit multiplies its weight by $e^{b_i}$ and renormalizes.
Choosing
\begin{equation}\label{eq:logit}
b_i = \begin{cases} \beta \log \tilde{w}_i + c, & i \in \mathcal{T}, \\ 0, & i \notin \mathcal{T}, \end{cases}
\qquad
c = \log \frac{m}{\sum_{k \in \mathcal{T}} a_k\, \tilde{w}_k^{\,\beta}}
\end{equation}
makes the denominator $e^{c} \sum_{k \in \mathcal{T}} a_k \tilde{w}_k^{\,\beta} + \sum_{k \notin \mathcal{T}} a_k = m + (1 - m) = 1$, so each text token receives $e^{c} a_i \tilde{w}_i^{\,\beta} = m\, r'_i$ by \eqref{eq:relay} and every other position keeps $a_i$: the result is exactly the relayed row.
The term $\beta \log \tilde{w}$ acts as a soft attention mask read from the LLM, and a token with $\tilde{w}_i = 0$ is masked outright.
The constant $c$ is the same for every text token in a head and holds the head's attention on the text at $m$; without it the bias would move attention from the text to the positions outside it, since $\tilde{w}_i \le 1$.

\paragraph{Alternative: the plain bias.}
We call \eqref{eq:logit} the \emph{mass-preserving bias}.
The \emph{plain bias} sets $c = \beta \log |\mathcal{T}|$ in every head instead, so that uniform weights change nothing: it multiplies a head's attention on the text by $(|\mathcal{T}|\,\tilde{w}_i)^{\beta}$ and renormalizes the whole row, where the mass-preserving bias renormalizes within the text.
Both give each head the same distribution $r'$ over the text and differ only in how much of its attention the head keeps there: under the plain bias, the odds $m/(1-m)$ are multiplied by
\begin{equation}\label{eq:kappa}
\kappa \;=\; |\mathcal{T}|^{\beta} \sum_{i \in \mathcal{T}} r_i\, \tilde{w}_i^{\,\beta},
\end{equation}
which is 1 for uniform $\tilde{w}$, above 1 when the head already attends to the tokens the LLM weights, and below 1 when it attends to others.
A head that agrees with the LLM then writes more from the text, and one that disagrees moves attention outside the text and writes less from it.
The mass-preserving bias leaves how much each head writes to the embedder, so that only a distribution over the text passes from the LLM.

\subsection{Why attention}\label{sec:why}

\paragraph{Only the weights can cross.}
Of what an instruction-tuned LLM computes while it reads a text under an instruction, an embedder could be handed either vectors or weights.
Vectors, whether hidden states, values or the representation of an answer, live in the LLM's own coordinates, with their own dimension and geometry, and cannot enter another model without a learned map.
Weights over the text have no coordinates: they are a distribution over the text itself.

\paragraph{The weights carry the instruction.}
They are also where the instruction acts.
In a Transformer, attention is the only operation that moves information between positions, and with the text before the instruction, the text's values are the same under every instruction.
The text's share of a head's output at the answer position, $\sum_{j=1}^{n} \alpha_j(x, q)\, v_j(x)$, therefore depends on the instruction only through its weights, so passing the weights passes everything the instruction changes in how the LLM reads the text, and nothing else.

\paragraph{The embedder already reads through weights.}
An embedder pools its text with a distribution over its tokens: uniform in mean pooling, its pooling token's attention in CLS or last-token pooling.
Once the two tokenizations are aligned through characters, the LLM's weights mean the same thing in the embedder, how much each part of the text matters for the question, and they enter the embedder where its own weights already are.
Which parts of a text matter for a question is a property of the text and the question, not of the model that reads them; how to represent those parts stays with the embedder.

\section{Experiments}\label{sec:exp}

We evaluate \method{} across LLMs, embedders and kinds of instruction, then change one ingredient at a time and ask what the relay passes, what it can select and what it changes.

\subsection{Setup}\label{sec:setup}

\paragraph{Models.}
We take the reading weights of six instruction-tuned LLMs: Qwen3 at 0.6B, 1.7B, 4B and 8B \citep{yang2025qwen3}, Llama-3.1-8B-Instruct \citep{grattafiori2024llama} and OLMo-3-7B-Instruct \citep{olmo2025olmo3}.
Four of their base checkpoints serve as controls.
We relay the weights into ten embedders that take no instruction: Qwen3-Embedding in its document mode \citep{zhang2025qwen3embedding}, BGE \citep{xiao2024cpack}, GTE \citep{li2023gte}, E5 \citep{wang2022e5} and two sentence-transformers models \citep{reimers2019sbert} (Appendix~\ref{app:settings}).
They span tokenizers, sizes and three kinds of pooling: the last token, [CLS] and the mean.
We read each LLM at one block, chosen without labels where its weights move most with the question, and keep one setting per pair on every task.
Unless stated otherwise, we relay Qwen3-1.7B's weights.

\paragraph{Tasks.}
We use InBedder's evaluation sets \citep{peng2024inbedder}.
NYT holds 2,144 news articles, labeled by topic and by the country where the news happened.
We embed every article once under the topic question and once under the location question, cluster each set of embeddings by the aspect its question asks about, and report the harmonic mean of the two V-measures.
IntentEmotion (IE) holds customer messages, labeled by how the customer feels and by what the customer needs, and arranges them in triplets.
Each triplet asks whether a message is closer to one that shares its emotion or one that shares its intent; an embedding that ignores the question scores 100, and one that always follows it scores 200.
Four more sets carry instructions of other kinds.
FewRel, FewNerd and FewEvent ask every text for the type of a marked relation, entity or event, and InstructSTSB gives every pair of sentences its own instruction.
We run them with Qwen3-1.7B here and with Llama and OLMo in Appendix~\ref{app:breadth}.
Every gain comes with a 95\% paired bootstrap interval (Appendix~\ref{app:metrics}).

\subsection{Main results}\label{sec:main}

\begin{table}[t]
\centering
\small
\caption{Five embedders that take no instruction, alone and relayed from Qwen3-1.7B (+\,relay), and Qwen3-Embedding-0.6B trained with instructions (bottom row). IE is 100 when the question is ignored and 200 when it is always followed.}
\label{tab:main}
\fontsize{8}{9.5}\selectfont
\setlength{\tabcolsep}{4pt}
\begin{tabular}{lcccccc}
\toprule
 & \multicolumn{4}{c}{Clustering, V-measure $\uparrow$} & STS, Spearman $\uparrow$ & Triplets $\uparrow$ \\
\cmidrule(lr){2-5}\cmidrule(lr){6-6}\cmidrule(lr){7-7}
Embedder & NYT & FewRel & FewNerd & FewEvent & InstructSTSB & IE \\
\midrule
Qwen3-Embedding-0.6B (document mode) & 55.9 & 54.4 & 42.7 & 59.1 & 0.0 & 100.0 \\
\quad + relay & 64.6 & 57.6 & 49.5 & 63.5 & 18.2 & 142.6 \\
\quad Gain & \textcolor{gainblue}{+8.7} & \textcolor{gainblue}{+3.2} & \textcolor{gainblue}{+6.8} & \textcolor{gainblue}{+4.5} & \textcolor{gainblue}{+18.2} & \textcolor{gainblue}{+42.6} \\
\midrule
bge-large-en-v1.5 & 56.7 & 55.4 & 43.3 & 59.0 & 0.0 & 100.0 \\
\quad + relay & 67.2 & 58.8 & 48.0 & 64.3 & 20.3 & 157.8 \\
\quad Gain & \textcolor{gainblue}{+10.6} & \textcolor{gainblue}{+3.4} & \textcolor{gainblue}{+4.7} & \textcolor{gainblue}{+5.3} & \textcolor{gainblue}{+20.3} & \textcolor{gainblue}{+57.8} \\
\midrule
gte-large-en-v1.5 & 56.8 & 54.1 & 42.3 & 56.8 & 0.0 & 100.0 \\
\quad + relay & 68.8 & 58.4 & 45.8 & 61.1 & 17.9 & 142.3 \\
\quad Gain & \textcolor{gainblue}{+12.0} & \textcolor{gainblue}{+4.3} & \textcolor{gainblue}{+3.5} & \textcolor{gainblue}{+4.3} & \textcolor{gainblue}{+17.9} & \textcolor{gainblue}{+42.3} \\
\midrule
all-mpnet-base-v2 & 56.7 & 52.1 & 40.9 & 55.5 & 0.0 & 100.0 \\
\quad + relay & 66.3 & 55.8 & 45.9 & 61.7 & 11.6 & 135.6 \\
\quad Gain & \textcolor{gainblue}{+9.6} & \textcolor{gainblue}{+3.7} & \textcolor{gainblue}{+5.0} & \textcolor{gainblue}{+6.2} & \textcolor{gainblue}{+11.6} & \textcolor{gainblue}{+35.6} \\
\midrule
e5-large-v2 & 54.3 & 56.4 & 41.4 & 61.1 & 0.0 & 100.0 \\
\quad + relay & 67.1 & 57.7 & 43.2 & 64.0 & 7.4 & 125.3 \\
\quad Gain & \textcolor{gainblue}{+12.8} & \textcolor{gainblue}{+1.2} & \textcolor{gainblue}{+1.8} & \textcolor{gainblue}{+2.8} & \textcolor{gainblue}{+7.4} & \textcolor{gainblue}{+25.3} \\
\midrule
Qwen3-Embedding-0.6B (instruction mode) & 68.4 & 63.4 & 58.1 & 67.3 & $-$1.7 & 190.0 \\
\bottomrule
\end{tabular}

\end{table}

\begin{figure}[t]
\centering
\includegraphics[width=\linewidth]{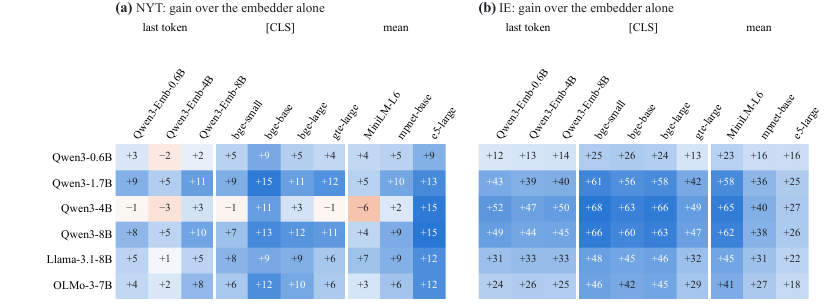}
\caption{The gain of relaying for every pair of an instruction-tuned LLM and an embedder.}
\label{fig:gains}
\end{figure}

Table~\ref{tab:main} and Figure~\ref{fig:gains} show the main result: embedders that never see the question now embed a text according to it.
Every embedder in Table~\ref{tab:main} gains on NYT and IE, and nearly every pair of an LLM and an embedder gains in Figure~\ref{fig:gains}.
The relay also raises the scores on FewRel, FewNerd, FewEvent and InstructSTSB, where the instruction does not pick one of two aspects.
Only e5-large-v2's gains on FewRel and FewEvent have intervals that include zero (Appendix~\ref{app:metrics}, Table~\ref{tab:main-ci}).
Qwen3-Embedding, trained with instructions and given the task, still scores higher on IE and the type sets but falls slightly below zero on InstructSTSB.
We speculate that instruction training, in which one instruction serves a whole task, teaches an embedder to shift every embedding of the task rather than to pick out the words a question asks about.

\subsection{Ablations}\label{sec:ablations}

\begin{figure}[t]
\centering
\includegraphics[width=\linewidth]{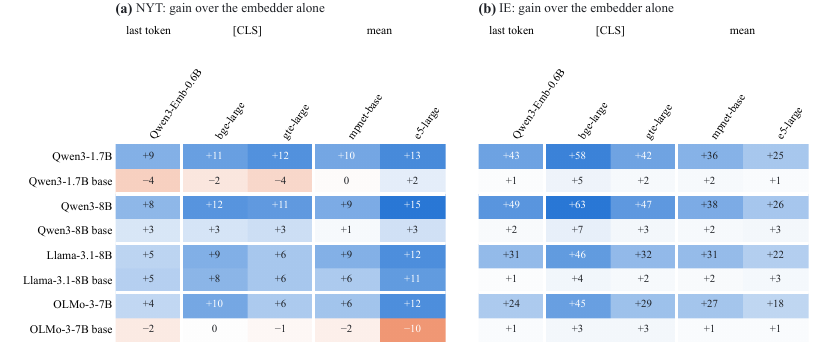}
\caption{The gain of relaying from each instruction-tuned LLM and from its base checkpoint.}
\label{fig:base}
\end{figure}

\paragraph{On instruction tuning.}
Base checkpoints relay much less into the five embedders of Table~\ref{tab:main} than their instruction-tuned siblings do (Figure~\ref{fig:base}).
Read at their siblings' blocks, the weights of the four base checkpoints barely move with the question.
The one case where a base's weights do move, Llama's on NYT, is also the one where a base relays nearly as much as its sibling.

\begin{figure}[t]
\centering
\includegraphics[width=\linewidth]{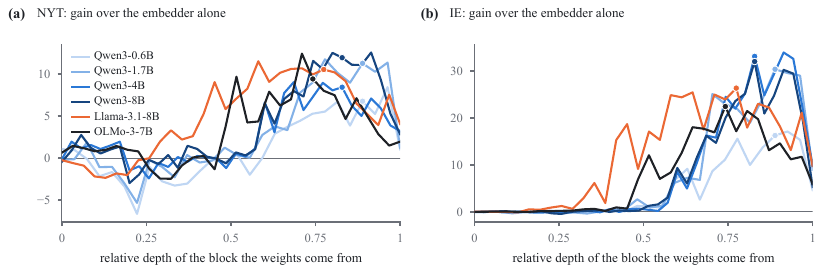}
\caption{The gain of relaying from each block of each LLM; dots mark the blocks we read.}
\label{fig:blocks}
\end{figure}

\paragraph{On the LLM block.}
The gain tracks the question sensitivity through the LLM's depth (Figures~\ref{fig:blocks} and~\ref{fig:premises}).
We relay the weights of every block of each LLM into the two embedders that pool by the mean, all-mpnet-base-v2 and e5-large-v2.
Blocks where the question has not yet moved the weights give nothing, and the gain appears where the weights start to move, earliest in Llama.
We can therefore choose the block without labels, as the one where the weights move most with the question; this choice loses little against the best block for every LLM but OLMo.

\paragraph{On the embedder's side.}
Relaying into the embedder's later layers gives nearly all of the gain, and into its earlier layers almost none (Appendix~\ref{app:layers}).
Replacing the mass-preserving bias with the plain bias moves the scores by about a point (Appendix~\ref{app:text-share}).

\subsection{Interpretability of Attention Relay}\label{sec:mechanism}

\begin{figure}[t]
\centering
\begin{minipage}[t]{0.47\linewidth}
\centering
\includegraphics[width=\linewidth]{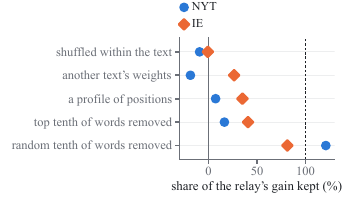}
\caption{The share of the gain kept when the weights or the text change (bge-large).}
\label{fig:controls}
\end{minipage}\hfill
\begin{minipage}[t]{0.47\linewidth}
\centering
\includegraphics[width=\linewidth]{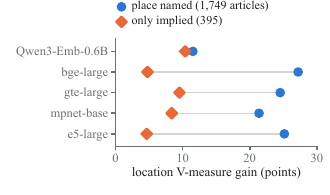}
\caption{The location gain where an article names the country and where it does not.}
\label{fig:split}
\end{minipage}
\end{figure}

\paragraph{What the relay passes.}
The gain comes from the words of each text that the LLM reads (Figure~\ref{fig:controls}).
Weights with the LLM's shape but the wrong words keep at most about a third of it, whether they are shuffled within the text, taken from another text or reduced to a profile of positions.
Removing the tenth of words the LLM weights most takes away most of the gain, while removing a random tenth takes at most a quarter.

\paragraph{What the relay can select.}
The relay gains more when the text names what the question asks about.
We split NYT's articles into those that name one of the ten countries or its people and those that do not, and cluster each set on its own.
Where an article names the country, the location gain is two to six times larger than where the place is only implied, for four of the five embedders (Figure~\ref{fig:split}).
Qwen3-Embedding, the one embedder built on an LLM, gains about as much in both.

\begin{figure}[t]
\centering
\includegraphics[width=\linewidth]{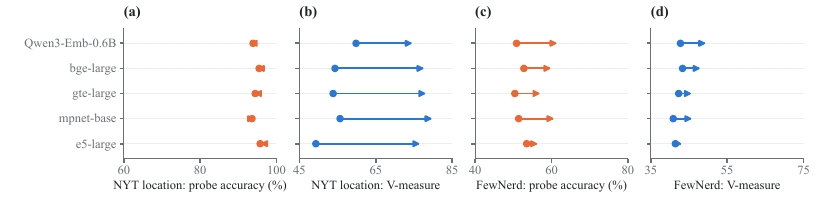}
\caption{Probe accuracy and V-measure of the asked aspect, from each embedder alone to relayed.}
\label{fig:probe}
\end{figure}

\paragraph{What the relay changes.}
Relaying makes the asked aspect count in one of two ways: it surfaces an aspect the embedding held but did not cluster by, or makes an aspect more readable in the first place.
A linear probe reads whether an embedding holds an aspect at all, and the V-measure whether the embedding clusters by it.
An NYT article runs to hundreds of words, and most articles name the country or its people.
Every embedding alone already holds the location, which a probe reads with 94 to 96\% accuracy; relaying leaves the probe within a point but raises location V-measure by 15 to 27 points.
A FewNerd text is one sentence of about 30 words, and its label is the type of one marked entity, often among entities of other types.
Pooled over the whole sentence, that entity takes a small share of the embedding; relaying moves the pooling onto it and raises the probe's accuracy by 3 to 11 points. Figure~\ref{fig:probe} shows the difference under these two scenarios.

\section{Related Work}\label{sec:related}

\paragraph{Instruction-aware text embeddings.}
A text can be grouped or compared in several valid ways, as studied in alternative clustering \citep{gondek2004nonredundant,niu2014iterative} and conditional similarity \citep{veit2017conditional}, as in C-STS \citep{deshpande2023csts}.
Instruction-aware embedders let an instruction choose the way and learn to follow it in training: InstructOR \citep{su2023instructor}, E5-Mistral \citep{wang2024improving}, GritLM \citep{muennighoff2024gritlm} and Qwen3-Embedding \citep{zhang2025qwen3embedding} through contrastive training with instructions, InBedder by learning to answer a question about the text \citep{peng2024inbedder}, and GSTransform through a transformation of a generic embedder's outputs, trained for each instruction \citep{feng2025gstransform}.
Even so, retrieval models trained with instructions use them mostly as keywords \citep[FollowIR;][]{weller2024followir}.
\method{} trains nothing: the instruction following comes from an LLM at inference, and the embedder is used as released.

\paragraph{Training-free embeddings from LLMs.}
An LLM's hidden state at the end of a prompt that asks for a text in one word serves as the text's embedding \citep[PromptEOL;][]{jiang2024scaling}; later methods average several prompts \citep[MetaEOL;][]{lei2024metaeol} or generated rewrites \citep[GenEOL;][]{thirukovalluru2024geneol}, repeat the text \citep[Echo embeddings;][]{springer2024repetition}, or pool the value vectors \citep[Value Aggregation;][]{zhang2026values}.
With a condition in the prompt, such readouts become conditional \citep[PonTE;][]{yamada2025ponte}, and self-contrastive steering improves them \citep[Focusing Condition;][]{cheng2026focusing}.
All of them represent the text in the LLM's own space; \method{} takes only the LLM's attention and keeps the space the embedder was trained for.

\paragraph{Attention as a question-conditioned signal.}
At the last position, attention heads fetch what a prediction needs from earlier tokens \citep{geva2023dissecting,wang2023ioi,wu2025retrieval}.
As a score, this attention re-ranks passages in ICR \citep{chen2025icr} and QRHead \citep{zhang2025qrhead}, ranks documents once calibrated for position \citep[Found in the Middle;][]{hsieh2024found}, decides which cached tokens to keep \citep[SnapKV;][]{li2024snapkv}, and, taken from a small model, which tokens a larger model reads \citep[SpecPrefill;][]{liu2025specprefill}.
Attention weights are not explanations in general \citep{jain2019attention,wiegreffe2019attention}, so we test the reading weights against shuffled, other-text and uniform weights.
These uses stay within one model or one tokenizer family; \method{} carries the weights through characters into a different model.

\paragraph{Changing attention at inference.}
Attention steering scales a model's attention on one span, marked by the user \citep[PASTA;][]{zhang2024tell}, found by the model \citep[AutoPASTA;][]{zhang2024autopasta} or holding the instruction, as in SpotLight \citep{venkateswaran2026spotlight} and InstABoost \citep{guardieiro2025instaboost}, by one factor.
Image editing, as in Prompt-to-Prompt \citep{hertz2023prompt} and Plug-and-Play \citep{tumanyan2023pnp}, injects the attention maps of one generation into another, which keeps its own values, and attention transfer \citep{zagoruyko2017attention,li2024attentiontransfer} and MiniLM \citep{wang2020minilm} train a student with a teacher's attention maps, also across tokenizers through character offsets \citep[MATT;][]{haltiuk2025matt}.

\section{Discussion}\label{sec:discussion}

\paragraph{It selects; it cannot add.}
Relaying changes only the text's share of a head's output.
At the pooling token, a head's output becomes
\begin{equation}\label{eq:relayed-head}
o' \;=\; m \sum_{i \in \mathcal{T}} r'_i\, v_i \;+\; \sum_{j \notin \mathcal{T}} a_j\, v_j ,
\end{equation}
with the embedder's own values and the same weight $m$ on the text; in mean pooling, the embedding is $\sum_i r'_i u_i$ over the embedder's own final states $u_i$.
Each relayed output is thus a convex combination of vectors the embedder computed itself: the instruction decides which of the embedder's content dominates, and only a distribution over the text, never a vector, passes from the LLM.
What the embedder does not compute cannot be relayed: knowledge the LLM draws on to answer, such as the country a city lies in, and anything an instruction asks beyond which tokens matter, such as how two entities relate.

\paragraph{A shared backbone.}
Many embedders built on an LLM keep its weights and add a small fine-tuning change, often a LoRA adapter \citep{hu2022lora}.
If the LLM's instruction tuning were an adapter too, one backbone could serve as both models: with one adapter it would read the text for the question, and with the other it would embed it.
The two would share their tokens, their heads and a single copy of the weights.
Modern post-training, however, changes every weight of the model rather than training an adapter.

\paragraph{Open questions.}
The relay needs attention whose weights form a distribution over the text, as softmax attention does.
Sparse attention, such as NSA \citep{yuan2025nsa}, reads only part of the text at full resolution, and linear attention, such as Gated DeltaNet \citep{yang2025gated}, keeps a state instead of weights; whether their implicit attention \citep{ali2025hidden} can be relayed, from such an LLM or into such an embedder, is open.
Nothing in the method is specific to text either: a vision-language model's attention over an image follows the question \citep{zhang2025mllms}, and aligning image patches by area instead of characters would relay it into an image encoder's pooling; whether this makes image embeddings instruction-aware, and embeddings of other modalities after them, is the next test.

\subsection*{AI use statement}
In this work, we used generative AI tools to write the code for the experiments and the analyses, to produce the figures and tables from the results files, to search and check the literature, and to draft sections of the paper.
The research questions and the central idea of the method came from the authors.
We have reviewed all AI-assisted work, and we take responsibility for the final content of this work, including text, claims or artifacts produced with the aid of generative AI.

\bibliography{refs}
\bibliographystyle{iclr2027_conference}

\clearpage
\appendix
\section{Implementation details}\label{app:impl}

\subsection{Settings}\label{app:settings}

Every number in the paper comes from one setting per pair of an LLM and an embedder, used unchanged on every task.
Tables~\ref{tab:models} and~\ref{tab:datasets} point to every model and dataset, and Table~\ref{tab:settings} gives each pair's setting.

\begin{table}[h]
\centering
\small
\caption{Every model in the paper.}
\label{tab:models}
\begin{tabular}{ll}
\toprule
Model & Hugging Face \\
\midrule
\multicolumn{2}{l}{\emph{Instruction-tuned LLMs}} \\
Qwen3-0.6B & \texttt{Qwen/Qwen3-0.6B} \\
Qwen3-1.7B & \texttt{Qwen/Qwen3-1.7B} \\
Qwen3-4B & \texttt{Qwen/Qwen3-4B} \\
Qwen3-8B & \texttt{Qwen/Qwen3-8B} \\
Llama-3.1-8B-Instruct & \texttt{meta-llama/Llama-3.1-8B-Instruct} \\
OLMo-3-7B-Instruct & \texttt{allenai/Olmo-3-7B-Instruct} \\
\addlinespace
\multicolumn{2}{l}{\emph{Base checkpoints}} \\
Qwen3-1.7B-Base & \texttt{Qwen/Qwen3-1.7B-Base} \\
Qwen3-8B-Base & \texttt{Qwen/Qwen3-8B-Base} \\
Llama-3.1-8B & \texttt{meta-llama/Llama-3.1-8B} \\
OLMo-3-7B & \texttt{allenai/Olmo-3-1025-7B} \\
\addlinespace
\multicolumn{2}{l}{\emph{Embedders, used without an instruction}} \\
Qwen3-Embedding-0.6B & \texttt{Qwen/Qwen3-Embedding-0.6B} \\
bge-large-en-v1.5 & \texttt{BAAI/bge-large-en-v1.5} \\
gte-large-en-v1.5 & \texttt{Alibaba-NLP/gte-large-en-v1.5} \\
all-mpnet-base-v2 & \texttt{sentence-transformers/all-mpnet-base-v2} \\
e5-large-v2 & \texttt{intfloat/e5-large-v2} \\
all-MiniLM-L6-v2 & \texttt{sentence-transformers/all-MiniLM-L6-v2} \\
bge-small-en-v1.5 & \texttt{BAAI/bge-small-en-v1.5} \\
bge-base-en-v1.5 & \texttt{BAAI/bge-base-en-v1.5} \\
Qwen3-Embedding-4B & \texttt{Qwen/Qwen3-Embedding-4B} \\
Qwen3-Embedding-8B & \texttt{Qwen/Qwen3-Embedding-8B} \\
\bottomrule
\end{tabular}

\end{table}

\begin{table}[h]
\centering
\small
\caption{Every dataset, from InBedder's evaluation \citep{peng2024inbedder}.}
\label{tab:datasets}
\begin{tabular}{ll}
\toprule
Dataset & Hugging Face \\
\midrule
\multicolumn{2}{l}{\emph{Used throughout}} \\
NYT & \texttt{BrandonZYW/NYTClustering} \\
IntentEmotion (IE) & \texttt{BrandonZYW/IntentEmotion} \\
\addlinespace
\multicolumn{2}{l}{\emph{Used in \S\ref{sec:exp}}} \\
FewRel & \texttt{BrandonZYW/FewRelClustering} \\
FewNerd & \texttt{BrandonZYW/FewNerdClustering} \\
FewEvent & \texttt{BrandonZYW/FewEventClustering} \\
InstructSTSB & \texttt{BrandonZYW/InstructSTSB} \\
\bottomrule
\end{tabular}

\end{table}

\paragraph{The LLMs.}
Each text is cut to its first 400 tokens in the LLM's tokenizer and given to the LLM in its chat template (Appendix~\ref{app:prompt}); a base checkpoint is given its instruct sibling's template.
The reading weights are taken at the block of Table~\ref{tab:llms}, counting from 0: the block with the highest question sensitivity (\S\ref{sec:motivation-llm}) in the second half of the network, excluding the last block, with the sensitivity averaged over the texts of NYT and IE and no labels read.
A base checkpoint is read at its sibling's block.

\begin{table}[h]
\centering
\small
\caption{Each LLM, the block its reading weights are taken at.}
\label{tab:llms}
\begin{tabular}{lll}
\toprule
LLM & Blocks & Block read \\
\midrule
Qwen3-0.6B & 28 & 24 \\
Qwen3-1.7B & 28 & 24 \\
Qwen3-4B & 36 & 29 \\
Qwen3-8B & 36 & 29 \\
Llama-3.1-8B-Instruct & 32 & 24 \\
OLMo-3-7B-Instruct & 32 & 23 \\
Qwen3-1.7B-Base & 28 & 24 \\
Qwen3-8B-Base & 36 & 29 \\
Llama-3.1-8B & 32 & 24 \\
OLMo-3-7B & 32 & 23 \\
\bottomrule
\end{tabular}

\end{table}

\paragraph{The embedders.}
Each embedder reads the same text, cut as above, and no instruction.
All relay the LLM's weights in the multiply form with $\beta = 1$ (\eqref{eq:relay}), and those that pool with attention do so with the mass-preserving bias in every head of every layer.
The weights reach the embedder's tokens through characters (\eqref{eq:align}); the first token of Qwen3-Embedding, the attention sink of its document mode, is counted outside the text.
e5-large-v2 reads every text after the prefix \texttt{"query:~"}, the input format its model card prescribes for clustering and classification; like the special tokens, the prefix lies outside the text and gets no relayed weight.

\begin{table}[h]
\centering
\small
\caption{The setting of each embedder, the same with every LLM.}
\label{tab:settings}
\begin{tabular}{llll}
\toprule
Embedder & Pooling & Relayed into & Max.\ tokens \\
\midrule
Qwen3-Embedding-0.6B & last token & attention & -- \\
Qwen3-Embedding-4B & last token & attention & -- \\
Qwen3-Embedding-8B & last token & attention & -- \\
bge-large-en-v1.5 & [CLS] & attention & 512 \\
bge-small-en-v1.5 & [CLS] & attention & 512 \\
bge-base-en-v1.5 & [CLS] & attention & 512 \\
gte-large-en-v1.5 & [CLS] & attention & 512 \\
all-mpnet-base-v2 & mean & pooling weights & 512 \\
e5-large-v2 & mean & pooling weights & 512 \\
all-MiniLM-L6-v2 & mean & pooling weights & 512 \\
\bottomrule
\end{tabular}

\end{table}

\FloatBarrier
\subsection{Chat template}\label{app:prompt}

Every LLM reads every text in its own chat template as released, with the template's default system message, if any, and thinking turned off where the template offers it.
The text and the task's instruction (Table~\ref{tab:instructions}) fill the user turn, and the answer position is the final token of the sequence.
For Qwen3, whose template is shown below, that token is the line break after \texttt{</think>}.

\begin{Verbatim}[frame=single, fontsize=\small, framesep=2mm, samepage=true]
<|im_start|>user
Text:
{text}

{instruction}

Answer in one word.<|im_end|>
<|im_start|>assistant
<think>

</think>

\end{Verbatim}

The sequence is tokenized piece by piece (the template up to \texttt{Text:}, the text, the instruction, the closing sentence and the template's end), so that no token crosses the text's boundaries.

\begin{table}[h]
\centering
\small
\caption{The instruction each dataset puts in the template's instruction slot.}
\label{tab:instructions}
\begin{tabular}{lp{0.72\linewidth}}
\toprule
Dataset & Instruction \\
\midrule
NYT, topic & What is the topic of news? \\
\addlinespace
NYT, location & Where did the news happen? \\
\addlinespace
IntentEmotion, emotion & How does the customer feel? \\
\addlinespace
IntentEmotion, intent & What does the customer need? \\
\addlinespace
FewRel & Here is a sentence. Please tell me the relation type between two specified entities appended after the sentence. \\
\addlinespace
FewNerd & Here is a sentence. Please tell me the type of the specified entity appended after the sentence. \\
\addlinespace
FewEvent & Here is a sentence. Please tell me the type of the specified event according to the trigger words appended after the sentence. \\
\addlinespace
InstructSTSB & each pair's own instruction, from the dataset \\
\bottomrule
\end{tabular}

\end{table}

\FloatBarrier
\subsection{Embedder inputs}\label{app:embedder-inputs}

Every embedder reads the text as its own tokenizer splits it, with its own special tokens.
Table~\ref{tab:embedder-inputs} gives what each embedder reads around the text in each condition, where \texttt{\{text\}} is the text, \texttt{\{question\}} the question or the set's instruction of Table~\ref{tab:instructions}, and \texttt{\{task\}} a task description of Table~\ref{tab:task-descriptions}.

\begin{table}[h]
\centering
\small
\caption{What each embedder reads around the text; \texttt{\textbackslash n} is a line break.}
\label{tab:embedder-inputs}
\begin{tabular}{>{\raggedright\arraybackslash}p{0.24\linewidth}>{\raggedright\arraybackslash}p{0.31\linewidth}>{\raggedright\arraybackslash}p{0.35\linewidth}}
\toprule
Embedder & Condition & Input \\
\midrule
Every embedder but e5-large-v2 & alone and relayed & \texttt{\{text\}} \\
\addlinespace
e5-large-v2 & alone and relayed & \texttt{query:~\{text\}} \\
\addlinespace
all-mpnet-base-v2, bge-large, gte-large & the question in the input, before or after the text (Figure~\ref{fig:question-input}) & \texttt{\{question\}~\{text\}}\newline \texttt{\{text\}~\{question\}} \\
\addlinespace
e5-large-v2 & the same & \texttt{query:~\{question\}~\{text\}}\newline \texttt{query:~\{text\}~\{question\}} \\
\addlinespace
bge-large, gte-large, all-mpnet-base-v2 & the set's instruction in the input (Table~\ref{tab:breadth-all}) & \texttt{\{question\}~\{text\}} \\
\addlinespace
e5-large-v2 & the same & \texttt{query:~\{question\}~\{text\}} \\
\addlinespace
Qwen3-Embedding-0.6B, instruction mode & the other sets' instructions (Tables~\ref{tab:main} and~\ref{tab:breadth-all}) & \texttt{Instruct:~\{question\}\textbackslash{}n}\newline \texttt{Query:\{text\}} \\
\addlinespace
Qwen3-Embedding-0.6B, instruction mode & NYT and IE (Table~\ref{tab:main} and Figure~\ref{fig:question-input}) & \texttt{Instruct:~\{task\}\textbackslash{}n}\newline \texttt{Query:\{text\}} \\
\bottomrule
\end{tabular}

\end{table}

\begin{table}[h]
\centering
\small
\caption{Task descriptions Qwen3-Embedding-0.6B reads in its instruction mode for NYT and IE.}
\label{tab:task-descriptions}
\begin{tabular}{ll}
\toprule
Question & Task description \\
\midrule
NYT, topic & Identify the topic of the given news article \\
NYT, location & Identify the location where the given news article happened \\
IE, emotion & Identify the emotion the customer expresses in the given message \\
IE, intent & Identify what the customer needs in the given message \\
\bottomrule
\end{tabular}

\end{table}

\FloatBarrier
\subsection{Controls}\label{app:controls}

Figure~\ref{fig:controls} asks whether the gain needs the LLM's weights on the text's own words.
Three controls keep the shape of the weights but cut their tie to the words.
\emph{Shuffled within the text} permutes the text's weights at random over its own tokens.
\emph{Another text's weights} gives the text, token by token, the weights of a random text of the same length, or the nearest length.
\emph{A profile of positions} gives each token the LLM's average weight at the same relative position, under the same question, over all the other texts, with relative position split into twenty bins.
It keeps where in a text the LLM tends to look, but not which words this text holds.
The shuffled and other-text controls average three random draws.
Two more controls remove words from each text before the embedder reads it: the tenth of its words that the LLM weights most under the question, or a random tenth.
On the shortened texts, we relay the LLM's weights on the remaining words and compare with the embedder alone on the same texts.

\section{Evaluation metrics}\label{app:metrics}

Let $e_q(x)$ be an embedder's embedding of the text $x$ under the question $q$; the embedder alone gives the same embedding under every question.

\paragraph{V-measure.}
For the class labels $Y$ and a clustering $K$ of the same texts \citep{rosenberg2007vmeasure},
\begin{equation}\label{eq:vmeasure}
h = 1 - \frac{H(Y \mid K)}{H(Y)}, \qquad
c = 1 - \frac{H(K \mid Y)}{H(K)}, \qquad
V = 100 \cdot \frac{2\,h\,c}{h + c},
\end{equation}
where $H$ is the entropy of the empirical distribution.
$K$ is $k$-means with as many clusters as classes, on the centered, unit-length embeddings, and $V$ is averaged over ten initializations.

\paragraph{NYT.}
With $V_{\mathrm{topic}}$ the V-measure of the topic labels on the embeddings under the topic question, and $V_{\mathrm{loc}}$ that of the country labels on the embeddings under the location question,
\begin{equation}\label{eq:nyt}
\mathrm{NYT} = \frac{2\, V_{\mathrm{topic}}\, V_{\mathrm{loc}}}{V_{\mathrm{topic}} + V_{\mathrm{loc}}}.
\end{equation}

\paragraph{IntentEmotion.}
For each question $q \in \{\mathrm{emotion}, \mathrm{intent}\}$, a triplet $(a, x^{+}, x^{-}) \in T_q$ holds an anchor message $a$, a message $x^{+}$ that shares only the anchor's answer to $q$, and a message $x^{-}$ that shares only its answer to the other question:
\begin{equation}\label{eq:ie}
\mathrm{IE} = 100 \sum_{q} \frac{1}{|T_q|} \sum_{(a, x^{+}, x^{-}) \in T_q} \mathbf{1}\big[\cos(e_q(a), e_q(x^{+})) > \cos(e_q(a), e_q(x^{-}))\big].
\end{equation}
The two sets hold the same messages with $x^{+}$ and $x^{-}$ swapped, so an embedding that ignores the question scores 100, and one that always follows it scores 200.
Figure~\ref{fig:question-input} reports $\mathrm{IE}/2$, the share of triplets that follow the question.
We use the triplets of the first 2,000 anchors in sorted order, 2,038 per question.

\paragraph{InBedder's other sets.}
FewRel, FewNerd and FewEvent report the V-measure of their type labels.
InstructSTSB reports $100\,\rho$, where $\rho$ is the Spearman correlation, over the sentence pairs $(s_1, s_2)$ with instruction $q$ and binary label $y$, between $\cos(e_q(s_1), e_q(s_2))$ and $y$.

\paragraph{Linear probe.}
The probe accuracy is the five-fold stratified cross-validated accuracy of a logistic regression ($C = 1$; \citealp{pedregosa2011scikit}) on the centered, unit-length embeddings; its chance level is the share of the largest class.

\paragraph{Gains and intervals.}
A gain is $\Delta = S_{\mathrm{relay}} - S_{\mathrm{alone}}$, a score of the relayed embedder minus the same score of the embedder alone.
The share of the gain that a control of Figure~\ref{fig:controls} keeps is $100\,\Delta_{\mathrm{control}} / \Delta$, where $\Delta_{\mathrm{control}}$ is the gain under the control (Appendix~\ref{app:controls}).
Each bootstrap replicate resamples the items (texts, triplets or sentence pairs) with replacement and scores both embeddings on the same resample, giving a replicate gain $\Delta^{*}$; the 95\% interval runs from the 2.5th to the 97.5th percentile of $\Delta^{*}$ over the replicates.
Table~\ref{tab:main-ci} gives every gain of Table~\ref{tab:main} with its interval.

\begin{table}[h]
\centering
\small
\caption{The gains of Table~\ref{tab:main}, with their 95\% paired bootstrap intervals below them.}
\label{tab:main-ci}
\begin{tabular}{lcccccc}
\toprule
Embedder & NYT & FewRel & FewNerd & FewEvent & InstructSTSB & IE \\
\midrule
Qwen3-Embedding-0.6B & +8.7 & +3.2 & +6.8 & +4.5 & +18.2 & +42.6 \\
 & {\scriptsize [2.9, 12.2]} & {\scriptsize [1.5, 5.0]} & {\scriptsize [4.4, 8.2]} & {\scriptsize [2.4, 6.4]} & {\scriptsize [16.6, 20.0]} & {\scriptsize [40.2, 45.1]} \\
\addlinespace
bge-large-en-v1.5 & +10.6 & +3.4 & +4.7 & +5.3 & +20.3 & +57.8 \\
 & {\scriptsize [7.5, 16.2]} & {\scriptsize [1.8, 5.0]} & {\scriptsize [2.1, 6.0]} & {\scriptsize [3.1, 6.7]} & {\scriptsize [18.5, 22.1]} & {\scriptsize [54.9, 60.6]} \\
\addlinespace
gte-large-en-v1.5 & +12.0 & +4.3 & +3.5 & +4.3 & +17.9 & +42.3 \\
 & {\scriptsize [6.7, 16.0]} & {\scriptsize [2.4, 5.8]} & {\scriptsize [1.3, 4.9]} & {\scriptsize [2.3, 5.7]} & {\scriptsize [16.3, 19.5]} & {\scriptsize [39.7, 44.9]} \\
\addlinespace
all-mpnet-base-v2 & +9.6 & +3.7 & +5.0 & +6.2 & +11.6 & +35.6 \\
 & {\scriptsize [6.0, 14.8]} & {\scriptsize [1.8, 4.5]} & {\scriptsize [2.8, 6.2]} & {\scriptsize [3.4, 7.8]} & {\scriptsize [10.6, 12.7]} & {\scriptsize [33.2, 37.9]} \\
\addlinespace
e5-large-v2 & +12.8 & +1.2 & +1.8 & +2.8 & +7.4 & +25.3 \\
 & {\scriptsize [8.8, 20.2]} & {\scriptsize [$-$0.6, 3.2]} & {\scriptsize [0.1, 3.7]} & {\scriptsize [$-$0.3, 5.6]} & {\scriptsize [6.7, 8.1]} & {\scriptsize [23.2, 27.5]} \\
\bottomrule
\end{tabular}

\end{table}

\FloatBarrier
\section{Additional experiments}\label{app:experiments}

\subsection{InBedder's other sets with other LLMs}\label{app:breadth}

Table~\ref{tab:breadth-all} repeats InBedder's other sets with the weights of Llama-3.1-8B-Instruct and OLMo-3-7B-Instruct beside Qwen3-1.7B's, and adds each embedder given the instruction in its input.\footnote{The embedder alone and the embedder given the instruction read each text as cut at Qwen3-1.7B's 400th token; Llama's and OLMo's cuts differ in 2 FewNerd and 6 FewEvent texts.}
OLMo's weights give the same picture as Qwen3-1.7B's; with Llama's, FewRel falls for every embedder.
Given the instruction in its input, every embedder scores below zero on InstructSTSB.

\begin{table}[h]
\centering
\small
\caption{InBedder's other sets with the weights of Qwen3-1.7B, Llama-3.1-8B-Instruct and
OLMo-3-7B-Instruct: each embedder alone, relayed, and given the instruction in its input.}
\label{tab:breadth-all}
\fontsize{8}{9.5}\selectfont
\begin{tabular}{lcccc}
\toprule
Embedder & FewRel & FewNerd & FewEvent & InstructSTSB \\
\midrule
\multicolumn{5}{c}{\emph{Qwen3-1.7B}} \\
\midrule
Qwen3-Embedding-0.6B & 54.4 $\to$ 57.6 & 42.7 $\to$ 49.5 & 59.1 $\to$ 63.5 & 0.0 $\to$ 18.2 \\
bge-large-en-v1.5 & 55.4 $\to$ 58.8 & 43.3 $\to$ 48.0 & 59.0 $\to$ 64.3 & 0.0 $\to$ 20.3 \\
gte-large-en-v1.5 & 54.1 $\to$ 58.4 & 42.3 $\to$ 45.8 & 56.8 $\to$ 61.1 & 0.0 $\to$ 17.9 \\
all-mpnet-base-v2 & 52.1 $\to$ 55.8 & 40.9 $\to$ 45.9 & 55.5 $\to$ 61.7 & 0.0 $\to$ 11.6 \\
e5-large-v2 & 56.4 $\to$ 57.7 & 41.4 $\to$ 43.2 & 61.1 $\to$ 64.0 & 0.0 $\to$ 7.4 \\
\midrule
\multicolumn{5}{c}{\emph{Llama-3.1-8B-Instruct}} \\
\midrule
Qwen3-Embedding-0.6B & 54.4 $\to$ 51.7 & 42.7 $\to$ 46.7 & 59.1 $\to$ 61.7 & 0.0 $\to$ 21.9 \\
bge-large-en-v1.5 & 55.4 $\to$ 52.0 & 43.3 $\to$ 44.4 & 59.0 $\to$ 62.7 & 0.0 $\to$ 23.0 \\
gte-large-en-v1.5 & 54.1 $\to$ 47.9 & 42.3 $\to$ 41.2 & 56.8 $\to$ 59.5 & 0.0 $\to$ 20.7 \\
all-mpnet-base-v2 & 52.1 $\to$ 50.9 & 40.9 $\to$ 41.8 & 55.5 $\to$ 58.5 & 0.0 $\to$ 14.1 \\
e5-large-v2 & 56.4 $\to$ 55.7 & 41.4 $\to$ 41.7 & 61.1 $\to$ 62.8 & 0.0 $\to$ 8.9 \\
\midrule
\multicolumn{5}{c}{\emph{OLMo-3-7B-Instruct}} \\
\midrule
Qwen3-Embedding-0.6B & 54.4 $\to$ 58.9 & 42.7 $\to$ 50.7 & 59.1 $\to$ 60.8 & 0.0 $\to$ 24.5 \\
bge-large-en-v1.5 & 55.4 $\to$ 59.3 & 43.3 $\to$ 50.1 & 59.0 $\to$ 61.3 & 0.0 $\to$ 25.2 \\
gte-large-en-v1.5 & 54.1 $\to$ 59.1 & 42.3 $\to$ 47.7 & 56.8 $\to$ 59.4 & 0.0 $\to$ 23.2 \\
all-mpnet-base-v2 & 52.1 $\to$ 56.6 & 40.9 $\to$ 48.1 & 55.5 $\to$ 59.1 & 0.0 $\to$ 15.8 \\
e5-large-v2 & 56.4 $\to$ 57.9 & 41.4 $\to$ 44.2 & 61.1 $\to$ 63.0 & 0.0 $\to$ 9.9 \\
\midrule
\multicolumn{5}{c}{\emph{Given the instruction in its input (Qwen3-Embedding: its instruction mode)}} \\
\midrule
Qwen3-Embedding-0.6B & 63.4 & 58.1 & 67.3 & $-$1.7 \\
bge-large-en-v1.5 & 56.1 & 44.6 & 61.5 & $-$13.7 \\
gte-large-en-v1.5 & 53.0 & 43.7 & 60.4 & $-$8.5 \\
all-mpnet-base-v2 & 56.2 & 46.9 & 64.9 & $-$11.2 \\
e5-large-v2 & 60.3 & 45.8 & 68.0 & $-$11.8 \\
\bottomrule
\end{tabular}

\end{table}

\subsection{On the embedder's layers}\label{app:layers}

\begin{figure}[h]
\centering
\begin{minipage}[t]{0.47\linewidth}
\centering
\includegraphics[width=\linewidth]{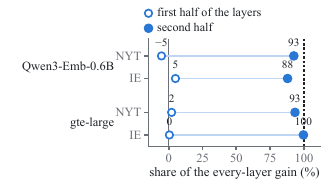}
\caption{Relaying into half of the embedder's layers, as a share of the every-layer gain.}
\label{fig:layers}
\end{minipage}\hfill
\begin{minipage}[t]{0.47\linewidth}
\centering
\includegraphics[width=\linewidth]{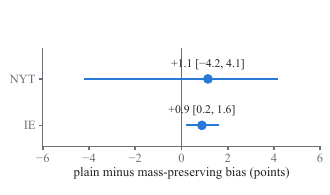}
\caption{The plain bias minus the mass-preserving bias, with its 95\% interval (bge-large).}
\label{fig:bias}
\end{minipage}
\end{figure}

The embedding keeps only what the embedder's later layers read.
Qwen3-Embedding-0.6B and gte-large pool with attention at every layer, so the weights can be relayed into some of their layers and not others: relayed into the second half alone, they give nearly the whole gain, and into the first half alone almost none (Figure~\ref{fig:layers}), although the pooling token attends to the text at least as much in the first half as in the second.
Averaged over the texts, the pooling token of the embedder alone puts 0.32 of its attention on the text in the first half of Qwen3-Embedding-0.6B's layers and 0.30 in the second on NYT (IE: 0.24 and 0.23), and 0.81 and 0.71 in gte-large's (IE: 0.53 and 0.42).

\subsection{On the text share}\label{app:text-share}

We compare the mass-preserving bias with the plain bias defined in \S\ref{sec:relay}.
What the LLM contributes is where each head reads within the text, not how much of its attention goes to the text.
The mass-preserving bias holds each head's attention on the text fixed, while the plain bias also lets it grow in the heads that agree with the LLM and shrink in the others; switching between them changes either score by about a point, a small fraction of the gain (Figure~\ref{fig:bias}).

\subsection{Cost}\label{app:cost}

Measured on one H100 at the runs' batch sizes, over 256 NYT articles under the location question, a full pass of Qwen3-1.7B costs 4.6 ms per article and question, and relaying adds at most 0.3 ms to the embedder's own pass, which takes 0.7 to 5.9 ms (Table~\ref{tab:cost}).

\begin{table}[h]
\centering
\small
\caption{Milliseconds per text and question on one H100: the full pass of Qwen3-1.7B, and each embedder alone and relayed.}
\label{tab:cost}
\begin{tabular}{lcc}
\toprule
Model & alone & relayed \\
\midrule
Qwen3-1.7B, the full pass & 4.58 & \\
\midrule
Qwen3-Embedding-0.6B & 2.28 & 2.61 \\
bge-large-en-v1.5 & 0.70 & 0.98 \\
gte-large-en-v1.5 & 1.13 & 1.40 \\
all-mpnet-base-v2 & 2.58 & 2.56 \\
e5-large-v2 & 5.88 & 5.99 \\
\bottomrule
\end{tabular}

\end{table}

\end{document}